# Improving GPS/INS Integration through Neural Networks

## Mathieu Nguyen-H and Chi Zhou

**Abstract**— The Global Positioning Systems (GPS) and Inertial Navigation System (INS) technology have attracted a considerable importance recently because of its large number of solutions serving both military as well as civilian applications. This paper aims to develop a more efficient and especially a faster method for processing the GPS signal in case of INS signal loss without losing the accuracy of the data. The conventional or usual method consists of processing data through a neural network and obtaining accurate positioning output data. The new or improved method adds selective filtering at the low-band frequency, the mid-band frequency and the high band frequency, before processing the GPS data through the neural network, so that the processing time is decreased significantly while the accuracy remains the same.

——————————— ◆ ———————————

## 1 INTRODUCTION

The Global Positioning System (GPS) is a satellite-based navigation system that provides reliable positioning, navigation, and timing services to worldwide users. Typically four or more satellites need to be tracked to estimate the position. GPS was originally intended for military applications, but in the 1980s, the government made the system available for civilian use. The GPS technology has attracted a considerable importance recently because of its large number of solutions serving both military as well as civilian applications. In order to provide optimum performance, the GPS technology needs to provide the greatest accuracy in positioning and also the smallest possible processing time for GPS data.

Most research on GPS technology has been focused on the improvement of the accuracy and the decrease of the noise influence. Actually the most popular and basic positioning technique is known as the GPS (Global Positioning System), however another technique which is quite widespread in the geo-referencing field is the INS (Inertial Navigation System) that uses sensors for getting positioning data. This paper aims to develop a more efficient and especially a faster method for processing the GPS signal in case of INS signal loss. Specifically, our main goal in this paper is to propose methods to decrease the processing time, of course, without losing the accuracy of the data. The conventional or usual method consists of processing data through a neural network and obtaining accurate positioning output data. The new or improved method adds selective filtering at the low-band frequency, the mid-band frequency and the high band frquency, before processing the GPS data through the neural network, so that the processing time is decreased significant-ly while the accuracy remains the same (measured in terms of the sum squared error).

As will be explained in the sequel, by processing only the frequency component needed by the user (and not all the components of the signal), the processing time can be significantly decreased. Moreover, this method may also be effetive in checking positioning data in case of dysfunction of a complementary back-up system. In this paper, we study the performance of such a designed system in terms of the processing time, convergence time and the mean-square-error of the output.

The paper is organized as follows. We provide technical background on the GPS and INS systems alogwith integrated GPS/INS in Section 2. We also present a brief description about the neural network based GPS/INS integration system. The system model for the proposed GPS data processing method is presented in Section 3. Section 4 presents simulation results followed by conclusions in Section 5.

## 2 BACKGROUND

This section provides background on the integrated GPS/INS system as well as application of the neural networks for the processing of GPS data.

### 2.1 GPS and INS Integrated System

The GPS is capable of providing accurate position information with respect to a navigation frame. The GPS system consists of a set of orbiting satellites whose locations are known and whose signals can be directly transmitted on the Earth. Only three satellites (with known positions) are necessary to determine with accuracy the position of the target [1]. The design of the actual system is such as that at least four satellites are always in direct line of sight (LoS) of every observatory point in the world

*M.Nguyen-H and C. Zhou are with the Electrical and Computer Engineering department, Illinois Institute of Technology, Chicago, IL.*

.

and the whole Global Positioning System is realized with 24 satellites distributed unevenly in symmetrically arranged orbital planes. Actually, GPS is an efficient system in that it can provide accurate position and velocity over long time period and can also offer service to an unlimited number of users anywhere on earth. Nevertheless, as this system relies on GPS satellite signals, it is susceptible to jamming, RF interference, multipaths and integrity problems. Moreover the GPS is not a self-contained autonomous system. In fact, the condition of direct line of sight of at least three or four satellites makes the functioning of the GPS difficult in urban area due to signal blockage and attenuation and can deteriorate the overall positioning accuracy. Furthermore, most of the error in GPS positioning come from medium propagation effects that are unpredictable to model.

On the other hand, the INS system involves a computer and a module containing gyroscopes and accelerometers or motion-sensing devices. It is a self-contained system integrating three acceleration components and three angular velocity components with respect of time. The purpose of INS is to deliver position, velocity and attitude components. Even if INS computes its own updated position and velocity by integrating information from the motion sensors, it requires initial position and velocity data from another external source such as GPS satellite receiver. The INS provides accurate position and velocity over short time periods (and also provide more accurate attitude information than the GPS does) but may slowly drift over short time [1]. Even if it is a self-contained system (ie it requires no external references to determine its position, its orientation or its velocity), the INS is not able to operate as a stand-alone navigation system. The residual bias error found in accelerometers and gyroscopes may deteriorate long-term positioning accuracy. So we can easily understand the INS suffers more form drifts rather than signal outages (which was the main concern of the GPS). However we have to be aware that the cost and the complexity of the INS place constraints on the environments in which the INS is in use.

One of the most used solutions is the INS/GPS integration [2]. This is the adequate solution to provide a navigation system that has superior performance in comparison with GPS or an INS stand-alone system. In fact, GPS/INS systems provide high position and velocity accuracy over short and longer time period. It gives navigational output during GPS signals outages and also high data rate, precise attitude determination and cycle slip detection and correction. As a matter of fact, GPS/INS integration compensates the inadequacies of the GPS and INS when utilized separately and provides a very efficient system

## 2.2 GPS/INS Systems Aided with Artificial Neural Network

Artificial neural network (ANN) is often designed for a specific application, such as pattern recognition or data classification, through a learning process. The main improvement of the neural networks with respect to Kalman filters is that there are adaptive training criterions in which neurons learn how to face new inputs.

One of the common type of ANN structure is to implement Radial Basis Function (RBF) with a simple architecture of only three layers (input, hidden and output layer) as described in Fig. 1 [3], [4, [5].

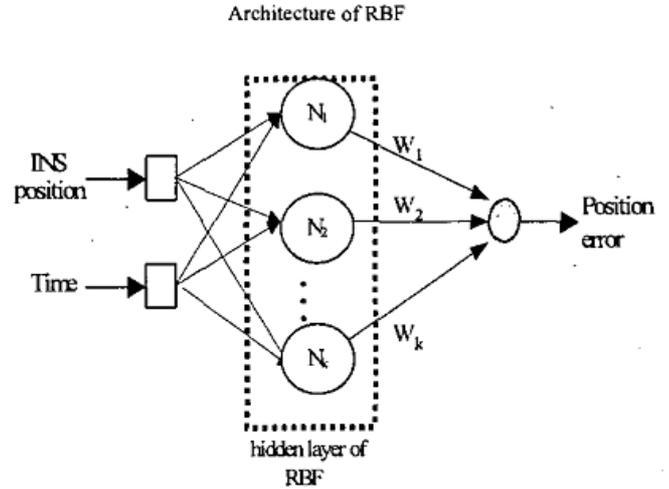

Figure 1. A RBF feed-forward neural network

Each hidden layer is basically a radial basis function characterizing the partitions of the input space and providing a degree of membership value for the input pattern with respect to the basis vector of the retrospective hidden unit itself. We can notice that the common basis functions are usually Gaussian functions and the output layer is built with linear neurons, which compute linear combination of the output of the hidden neurons. Moreover it has been shown that increasing the number of layer improves the precision of the system which also eases the observation and the adjustment of the training results [6], [7].

According to its simple structure, the RBF-ANN has faster training than the other networks [8]. The RBF-ANN procedure can be exlained as a two step procedure: In the first step, unsupervised fast training process determines the parameter of the hidden layer, while in the second step, this is a supervised training method (using the GPS/INS error provided by the wavelet multi-resolution analysis (WMRA) module) which determines the weight of the output layers. Practically, the error between the network output and the desired output is used to update the weights of the output layer and as the weights are defined linearly, the second stage of training is also fast. Finally, the GPS/INS navigation will mainly rely on GPS data, but in case of GPS outages, the INS position will be considered as the input of the RBF-ANN to predict the corresponding position information for each position components [9], [10].

## 3 SYSTEM MODEL

As seen in the previous sections, one of the more accurate ways to obtain accurate position on the Earth is to process GPS and INS data through artificial neural network. The goal of this paper is to run an RBF-ANN, study its characteristic and propose a new faster method to calculate positioning data without losing the whole accuracy especially in case of GPS outage or INS dysfunction. Depending on the application decided by the user, the suggested method will deal with low frequency, mid-frequency or high frequency component only. The proposed faster prediction method using RBF-ANN may have the following applications. 1) Neural networks are very useful in pattern recognition and in the field of the image compressing. Hence we could envision a system where the high frequency would represent the mobile to localize, the mid frequency as a troup of soldier to protect and the low frequency for the topography of the terrain. 2) Civil aopplications may also be envisioned. For instance, in the maritime field, for positioning one boat in the sea in the middle of the "boat traffic", the high frequency would again represent the mobile to localize, the middle one could be all obstacles the boat could meet and the low frequency could represent the topography of the terrain.

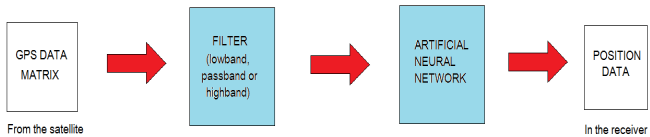

Figure 2. Block diagram of the proposed method.

As it can be seen from Fig. 2, the GPS data are stored under a matrix format, where each column represents a positioning component (North, East, altitude). Depending on the application, a filter is chosen to pre-process the incoming GPS signal. Supposing this is a lowband filter, the low frequency components are then processed in the neural network (i.e., through the teaching-learning process). By only processing the useful information, the users save time during the data processing; the remaining condition is to keep an acceptable accuracy in the output position data. So the proposed improved method or new method will stand for the model where the GPS data is pre-processed with a filter and then processed by the neural network; while the usual or conventional method will stand for the method where the GPS data are just processed by the neural network without pre-processing.

## 4 SIMULATION RESULTS

All the simulations are run under Matlab 7.0.0.190920 (R14) and with the toolbox *Constell*. We create a study environment where we consider the GPS signal as a matrix with one column for each component (North, East, and the altitude). The toolbox recreates the data sent by the satellites to a receiver on the surface of the Earth. In this approach, we consider a dysfunction of the INS acquisition and we process the GPS signal through the artificial neural network to decrease the noise and obtain more and more accurate information. Moreover, if we store the INS data under the same format, the role of the GPS and INS may be symmetric from a software point of view, and the method developed here may apply to the INS signal in case of GPS outage.

For the performance comparison between the conventional method and the proposed method, a GPS signal is created and is first processed through a neural network (conventional method). In a second experiment, the same GPS signal is pre-processed through filter and then processed thorugh neural network (improved method). For the proposed method, the action of filter is added: for the low frequency a lowband filter is simulated by picking the low frequency component over the three components constituting a GPS signal; the same is realized for a high-band component (a high-band filter is simulated) and a middle band component (where a band pass is simulated). In both cases, the simulations are run under the modifications of three parameters: the sum square error bound (error limit that the neural network has to support), the maximal size of neurons in the network, and the spread constant sigma value. Moreover, a clock has been implemented so that the time of computation of the neural network is measured. At the end the processing time of the neural network and the mean square error in output are stored as the output. In the following simulations the interest is focused on the comparison between the learned data and the original one and the time the whole process requires.

In each simulation result presented below, the blue curve represents the initial GPS data signal without any processing; the green dotted one represents the teaching part of the neural network, i.e., the part of information given by the neural network to produce an output as closed to the real GPS data as possible. And the purple line is the learned data ie the data learned from the processing (through neural network) of the noisy input (the GPS data with an added noise).

The six following figures are drawm in case of a low filter in the pre-processing stage. (i.e., the low frequency component is considered).

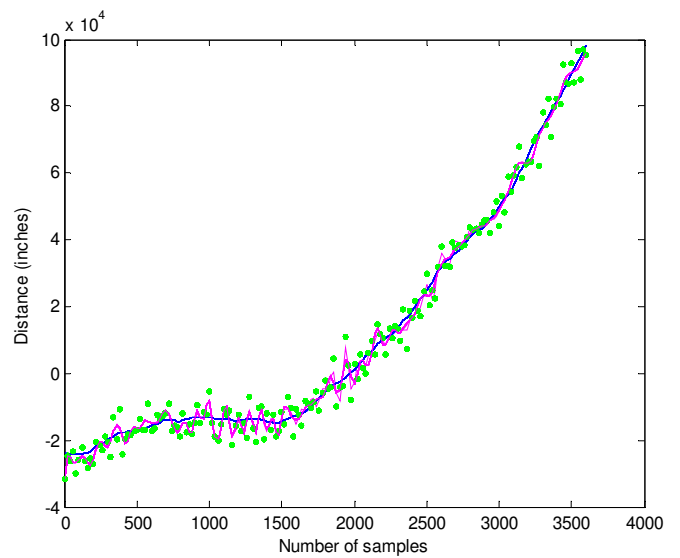

Fig. 3. Simulation for the conventional method

(nnsizze=100, sc=50, sum-squared error of 1x10^(-6))

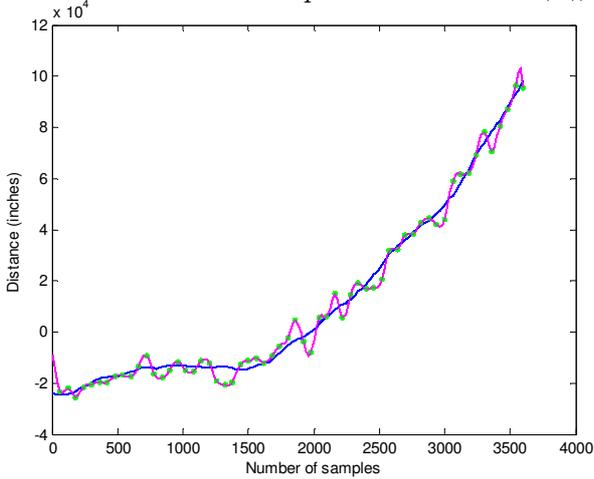

Fig. 4. Simulation for the improved method (nnsizze=100, sc=50, sum-squared error of 1x10^(-6))

It is seen from Figs. 3 and 4 that the accuracy is similar in both the conventional and the improved filtering method and also that it increases significantly after the time sample 2500. Before this value the measurement has no real meaning in term of accuracy, because the teaching process needs some initialization to be really efficient. Nevertheless, the systems reveal their own limit for too lower values: for sc=30, nnsize=50 and a sum-squared error of 1x10^(-6) as indicated in Figs. 5 and 6 below.

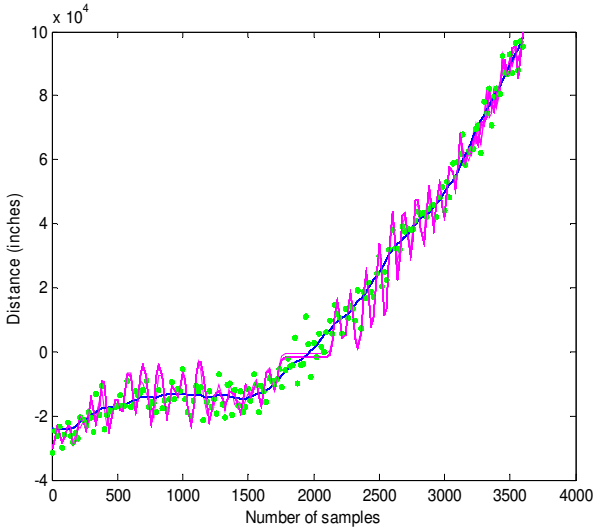

Fig. 5. Simulation for the conventional method (nnsizze=50, sc=30, sum-squared error of 1x10^(-6)) .

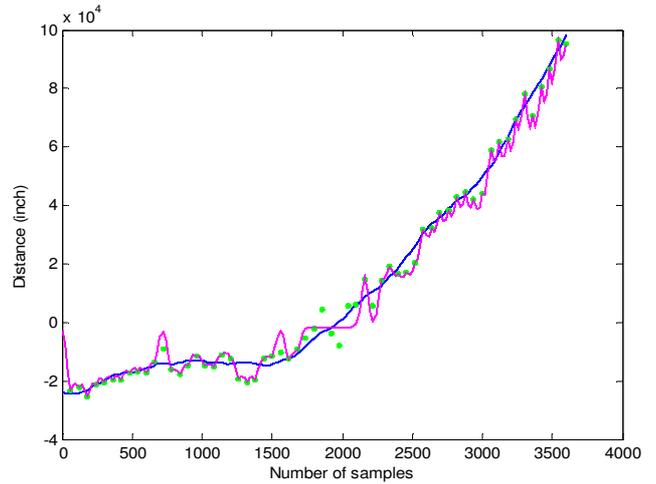

Fig. 6 Simulation for the improved method (nnsizze=30, sc=50, sum-squared error of 1x10^(-6))

It is observed from Figs. 5 and 6 that the data are not locally accurate because of the peaks in the curves.

Finally, we increase the sum-squared error parameter to 1x10^(-10). The accuracy of the method increases as it is shown in the next figure.

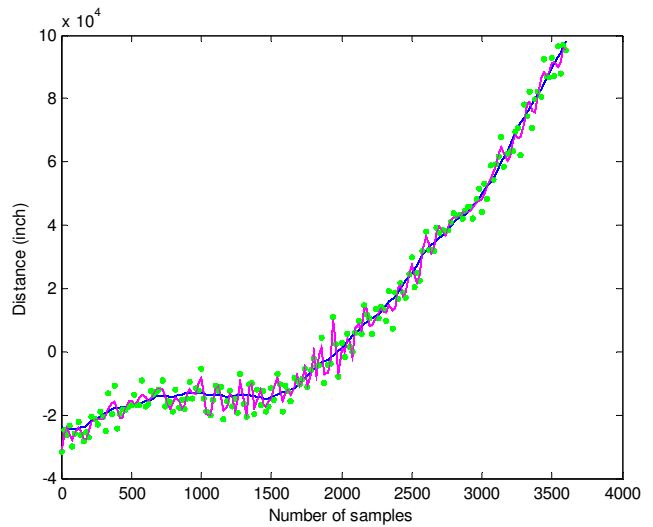

Fig. 7. Simulation for the conventional method (nnsizze=100, sc=100, sum-squared error of 1x10^(-10))

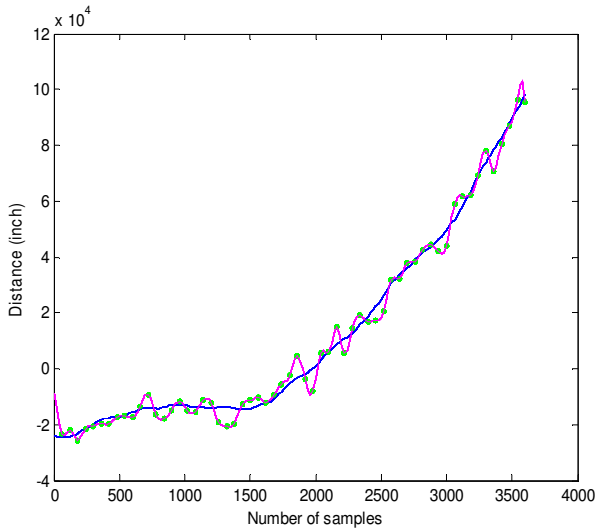

Fig. 8. Simulation for the improvedl method (nnsizze=100, sc=100, sum-squared error of 1x10^(-10)) .

Tables 1-4 below present the summary of simulation results when the sum squared error is 1x10^-6 and 1x10^-10.

Table 1. Simulation with sum squared error = 1x10^-6

| nnsize | 50 | 50 | 100 | 100 |
|---|---|---|---|---|
| Sc | 30 | 50 | 50 | 100 |
| Time-basic method | 1min 07sec | 1 min 08sec | 3min 14sec | 3min 14sec |
| MSE-basic method | 1.2228 *10^-24 | 1.2320 *10^-24 | 0.0108 | 0.0108 |
| Time-new method | 48.65 sec | 48.65sec | 51.3sec | 52sec |
| MSE-new method | 0.0681 | 0.0808 | 0.0808 | 0.900 |

Table 2. Simulation with sum squared error = 1x10^-10

| nnsize | 50 | 50 | 100 | 100 |
|---|---|---|---|---|
| Sc | 30 | 50 | 50 | 100 |
| Time-basic method | 1min 5sec | 1min 31 | 3min 13sec | 3min 25sec |
| MSE-basic method | 1.2228 *10^-24 | 1.2320 *10^-24 | 0.0108 | 0.0108 |
| Time-new method | - | 55.4sec | 49,85sec | 1min 37sec |
| MSE-new method | 0.0681 | 0.0808 | 0.0808 | 0.900 |

As we could expect, the processing time increases with the requirement on the sum squared error but the general trend shows a very significant improvement of the results when the new method is used (except for too small val-ues). In particular, the new proposed method improves the processing time with a factor between 2 and 3 and for every case the MSE still remains lower than 1.

Tablew 3 and 4 indicate simulation results when the pre-processing stage employes passband and highband filters. The results are presented for a sum squared error equal to 1x10^-10.

Table 3. Simulation with sum squared error = 1x10^-10 (with passband filter)

| nnsize | 50 | 50 | 100 | 100 |
|---|---|---|---|---|
| Sc | 30 | 50 | 50 | 100 |
| Time-basic method | 42.66sec | 1 min 5.6sec | 2min 22sec | 3min 14sec |
| MSE -basic method | 4.4384 *10^-34 | 1.436 *10^-24 | 1.73 *10^-20 | 1.68 *10^-15 |
| Time-new method | 32.1sec | 48.42sec | 50.45sec | 51sec |
| MSE-new method | 0.0849 | 0.0890 | 0.0867 | 0.0830 |

Table 4. Simulation with sum squared error = 1x10^-10 (with highband filter)

| nnsize | 50 | 50 | 100 | 100 |
|---|---|---|---|---|
| Sc | 30 | 50 | 50 | 100 |
| Time-basic method | 1min 6sec | 1 min 09sec | 2min 35sec | 2min 49ec |
| MSE-basic method | 1.1150 *10^-24 | 1.2149 *10^-24 | 9.6 *10^-20 | 1.46 *10^-20 |
| Time- new method | 48.1260 sec | 49.28sec | 50.74sec | 51sec |
| MSE -new method | 0.0637 | 0.0771 | 0.0781 | 0.8000 |

It can be seen that the passband and highband filtering in the pre-processing stage of GPS data improve the processing time considerably.

## 5 CONCLUSION

GPS/INS integration is a very efficient and wide-spread technique in every positioning problem but may sometimes suffer from GPS outages or INS dysfunction. The conventional or usual method of GPS data processsing consists of processing data through a neural network and obtaining accurate positioning output data. The proposed method adds selective filtering at the low-band frequency, the mid-band frequency and the high band frquency, before processing the GPS data through the neural network and offers a better solution in case of loss or degradation either in case of INS signal loss. So far, the new method has been able to improve the processing time significantly while maintaining the accuracy of the process.